\documentclass{article}
\usepackage{arxiv}

\usepackage{amsmath}

\usepackage{amssymb}
\usepackage{latexsym}

\usepackage{url}
\usepackage{xcolor}
\usepackage{color}

\usepackage{hyperref}
\usepackage{mathtools}

\definecolor{newcolor}{rgb}{.8,.349,.1}

\title{Multi Proxy Anchor Family Loss for Several Types of Gradients.}

\author{
	Shozo Saeki \\
	Center for Information Technology \\
	Ehime University \\
	Matsuyama, Ehime and 790-8577, Japan \\
	\texttt{saeki.shozo.cg@ehime-u.ac.jp} \\
	\And
	Minoru Kawahara \\
	Center for Information Technology \\
	Ehime University \\
	Matsuyama, Ehime and 790-8577, Japan \\
	\texttt{kawahara@ehime-u.ac.jp} \\
	\And
	Hirohisa Aman \\
	Center for Information Technology \\
	Ehime University \\
	Matsuyama, Ehime and 790-8577, Japan \\
	\texttt{aman@ehime-u.ac.jp} \\
}

\renewcommand{\shorttitle}{MPA Family Loss}

\begin{document}

\maketitle

\begin{abstract}
The deep metric learning (DML) objective is to learn a neural network that maps into an embedding space where similar data are near and dissimilar data are far.
However, conventional proxy-based losses for DML have two problems: gradient problem and application of the real-world dataset with multiple local centers.
Additionally, the performance metrics of DML also have some issues with stability and flexibility.
This paper proposes three multi-proxies anchor (MPA) family losses and a normalized discounted cumulative gain (nDCG@k) metric.
This paper makes three contributions.
(1) MPA-family losses can learn using a real-world dataset with multi-local centers.
(2) MPA-family losses improve the training capacity of a neural network owing to solving the gradient problem.
(3) MPA-family losses have data-wise or class-wise characteristics with respect to gradient generation.
Finally, we demonstrate the effectiveness of MPA-family losses, and MPA-family losses achieves higher accuracy on two datasets for fine-grained images.
\end{abstract}




\section{Introduction}
The deep metric learning (DML) objective is to learn a neural network that maps into an embedding space where similar data are near and dissimilar data are far.
Here, the similarities are computed using a specific metric, for example, L1 distance and L2 distance.
DML has been applied to few-shot learning \cite{FewShot}, face recognition \cite{Triplet}, and image retrieval tasks \cite{Cars, CUB, InShop}.
The DML results for these tasks have achieved state-of-the-art accuracy \cite{MS, XBM, SoftTriple}.

The DML loss function is the factor that greatly affects the accuracy.
Loss functions roughly classify two types of pair-based and proxy-based losses.
A proxy means the center of the data distribution for each class.
Then, proxy-based loss uses similarity between data and these proxies.
On the other hand, pair-based loss uses similarity between data.
The pair-based losses are significantly influenced by batch sampling \cite{Triplet, Npair, Contrastive, LiftedStructure, MS}. 
Conversely, the proxy-based losses are less affected by batch sampling \cite{ProxyNCA, SoftTriple, ProxyAnchor}.
Furthermore, a cross-batch memory module (XBM) \cite{XBM} and deep variational metric learning (DVML) \cite{DVML} are also proposed to improve the performance of DML.

ProxyNCA loss and ProxyAnchor loss, among proxy-based losses, have only one proxy for each class \cite{ProxyNCA, ProxyAnchor}.
However, the classes in practical datasets could have some local centers caused by intra-class variance, and one proxy cannot represent these structures \cite{SoftTriple, DMLSSR}.
In contrast, SoftTriple loss has multiple centers for each class to capture manifold structures \cite{SoftTriple}.
On the other hand, ProxyNCA loss and SoftTriple loss have similar properties to the softmax function \cite{ProxyNCA, SoftTriple}.

In this paper, we review the properties of SoftTriple loss and check gradients. 
Next, we propose the multi-proxies anchor (MPA) loss which extends from SoftTriple loss and ProxyAnchor loss.
MPA loss focuses on one problem for each ProxyAnchor loss and SoftTriple loss.
(1) MPA loss can learn the complex manifold embedding space because it considers multi-proxies compared with ProxyAnchor loss.
(2) MPA loss solves the gradient issues in SoftTriple loss by extending the loss function to like ProxyAnchor loss.
Furthermore, we focus on how the gradient of the MPA loss is generated.
ProxyAnchor loss and MPA loss are class-wise gradient generation, while SoftTriple loss is data-wise gradient generation.
We check the structure of MPA loss and review the gradient factors.
Finally, we propose data-wise multi-proxies anchor (MPA-DW) loss and data-wise and all-paired multi-proxies anchor (MPA-AP) loss.
MPA-DW loss and MPA-AP loss are more stable and have more training capacity because these losses are data-wise losses and have more gradient factors.
MPA loss, MPA-DW loss, and MPA-AP loss collectively refered to as MPA-family losses from now on

Furthermore, conventional performance metrics for DML tasks cannot fully evaluate performance and are unstable \cite{RealityCheck}.
Recently, the proposed MAP@R metric has been a stable performance metric \cite{RealityCheck}.
However, we assume that this metric cannot fully evaluate the performance in specific datasets. 
This paper checks the MAP@R metric properties and suggests a way to assess DML tasks using normalized discounted cumulative gain (nDCG), a more stable and flexible performance metric.

The contributions of this paper are as follows:
(1) MPA-family losses can learn the class distributions with multi-local centers in the embedding space compared to ProxyAnchor loss.
(2) MPA-family losses are more training capacity of a neural network than SoftTriple loss.
(3) MPA-family losses have data-wise or class-wise properties with respect to gradient generation.

An overview of the rest of the paper is as follows.
In section 2, we review the related works in this area.
Section 3 describes the MPA-family losses and nDCG@k metric for DML tasks.
Section 4 presents implementation details, a comparison of state-of-the-art losses, and validation of the number of proxies impacts.
Finally, section 5 concludes this work and discusses future works.

\section{Related Works}
In this section, we introduce loss functions for DML, which significantly affect the performance of DML.
DML loss functions are categorized into pair-based and proxy-based loss.
Following, we describe pair-based and proxy-based losses.

\subsection{Pair-based Losses}
The pair-based losses compute similarities between data in a feature space, and then losses are computed based on these similarities \cite{Contrastive,Triplet,LiftedStructure,Npair,HTL,MS}.
For example, Triplet loss \cite{Triplet} is computed using anchor, positive, and negative data.
Note that positive data are similar data against anchor data, and negative data are dissimilar data against anchor data.
Then similar data is of the same class as the anchor data, otherwise dissimilar data.
A combination of anchor and positive data is a positive pair, and a combination of anchor and negative data is a negative pair.
Triplet loss encourages larger a similarity of a positive pair than a negative pair \cite{Triplet}. 
Alternatively, lifted structured loss\cite{LiftedStructure}, N-pair loss \cite{Npair}, multi-similarity loss \cite{MS}, and the others compute losses for all pairs in a mini-batch.
These pair-based losses can redefine a general pair weighting (GPW) \cite{MS} framework that uses a unified weighting formulation.

However, DML learns by dividing the training data into mini-batches, such as a general neural network training framework.
Furthermore, the size of the datasets has recently increased incredibly.
Therefore, computing the loss for all combinations of training data is difficult.
For this reason, learning DML with pair-based losses is greatly affected by sampling mini-batch.
A good sampling strategy is crucial for good performance and fast convergence in a pair-based loss \cite{Triplet}.
However, good sampling is challenging, and learning results are easily variable. 
The cross-batch memory (XBM) \cite{XBM} module preserves the embeddings of the previous batch to learn the network using the target batch and the previous batches, where XBM assumes a "slow drift" phenomenon.
This module can use more learning pairs, despite the small memory cost.
However, this module must use an appropriate memory size; otherwise, the accuracy of the DML will decrease \cite{XBM}.

\subsection{Proxy-based Losses}
Proxy-based losses consider proxies in addition to training data and compute a loss with proxies.
The concepts of these losses mitigate the variability of the learning results due to the sampling strategy \cite{SoftTriple}.
The proxy-based losses are ProxyNCA loss \cite{ProxyNCA}, ProxyAnchor loss, and SoftTriple loss \cite{SoftTriple}.
Compared to pair-based losses, the SoftTriple loss and ProxyAnchor have state-of-the-art accuracy \cite{SoftTriple, ProxyAnchor}, even though the sampling strategy is random sampling.

The main differences between these losses are the number of proxies and the structure of the functions.
ProxyNCA loss and ProxyAnchor loss consider a single proxy for each class, and SoftTriple loss considers multiple proxy \cite{ProxyNCA, SoftTriple, ProxyAnchor}.
Single proxy loss cannot learn the complex manifold structure where class distributions in the embedding space have multiple local centers \cite{SoftTriple}.
In contrast, multiple proxies loss helps learn about complex manifold structures.
Furthermore, ProxyNCA loss has a gradient problem that affects backpropagation learning \cite{ProxyAnchor}.
ProxyAnchor loss improves the gradient problem of ProxyNCA loss \cite{ProxyAnchor}, and improves accuracy.
Note that SoftTriple loss and ProxyNCA loss are similar structures \cite{SoftTriple, ProxyNCA}.
	
\section{Proposed Loss Function and Evaluation Metric}
This section proposes multi-proxies anchor loss and a new DML evaluation metric for accurate comparison.
First, we review SoftTriple loss and check the gradient.
Next, we propose multi-proxies anchor loss, which extends SoftTriple loss and ProxyAnchor loss.
Then, we also propose data-wise multi-proxies anchor loss and all-paired multi-proxies anchor loss.
Finally, we also propose DML evaluation using nDCG, which is for a more accurate comparison.

\subsection{The Nature of SoftTriple Loss}
First, we discuss SoftTriple loss and introduce the notations used in this paper.

Let $X = \left\{ \mathbf{x}_{1}, \mathbf{x}_{2}, \cdots, \mathbf{x}_{N} \right\} \in \mathbb{R}^{N \times d}$ denote the feature vectors where $N$ is the number of data and $d$ is the dimension.
$c_{i}$ denote the corresponding label of data $\mathbf{x}_{i}$.
The proxy-based DML losses compute the similarity $\mathcal{S} \left( \mathbf{x}_{i}, c \right)$ between the instance $\mathbf{x}_{i}$ and the class $c$.
In SoftTriple loss, each class has multiple $K$ proxy centers; each center is denoted by $\mathbf{w}_{ck} \in \mathbb{R}^{d}$, where $c$ is a class index, and $k$ is a center index.
Generally, $\mathbf{x}_{i}$ and $\mathbf{w}_{ck}$ are normalized using L2 normalization; therefore, the L2 norm of $\mathbf{x}_{i}$ and $\mathbf{w}_{ck}$ equals one.
The similarity between data $\mathbf{x}_{i}$ and class $c$ is defined as \cite{SoftTriple}
\begin{equation}
	\label{original_similarity}
	\mathcal{S} \left( \mathbf{x}_{i}, c \right) = \sum_{k} \frac{\exp \left( \frac{1}{\gamma} \mathbf{x}_{i}^{\mathrm{T}} \mathbf{w}_{ck} \right)}{\sum_{l} \exp \left( \frac{1}{\gamma} \mathbf{x}_{i}^{\mathrm{T}} \mathbf{w}_{cl} \right)} \mathbf{x}_{i}^{\mathrm{T}} \mathbf{w}_{ck},
\end{equation}
where $\gamma$ is a hyperparameter.
The proxy-based DML losses mitigate batch sampling effects by computing the similarity using feature vectors of the data and proxy centers for the class.
Alternatively, in the pair-based DML losses, the similarity is computed by the dot product or Euclidean distance between the data \cite{Contrastive, Triplet, MS, XBM}.
Thus, compared to the similarity of the proxy-based losses, the similarity of the pair-based losses is highly dependent on the combination of data.

SoftTriple loss is the combination of the similarity loss and the regularization loss of the proxies \cite{SoftTriple}.
The similarity loss for data $\mathbf{x}_{i}$ is defined as
\begin{align}
	&L_{Sim} \left( \mathbf{x}_{i} \right) = -\log \frac{l^{+}_{ST} \left( \mathbf{x}_{i}, c_{i} \right)}{l^{+}_{ST} \left( \mathbf{x}_{i}, c_{i} \right) + \sum_{c^{\prime} \not= c_{i}} l^{-}_{ST} \left( \mathbf{x}_{i}, c^{\prime} \right)},
\end{align}
where $l^{+}_{ST} \left(\mathbf{x}, c \right) = \exp \left( \lambda \left( \mathcal{S} \left( \mathbf{x}, c \right) - \delta \right) \right)$ and $l^{-}_{ST} \left(\mathbf{x}, c \right) = \exp \left( \lambda \mathcal{S} \left( \mathbf{x}, c \right) \right)$.
$\delta$ denotes the margin, and $\lambda$ is the hyperparameter.
The center regularization loss minimizes the L2 distance between centers, and then the center regularization loss is defined as
\begin{equation}
	\label{CenterReg}
	L_{Regularizer} = \frac{\sum_{j}^{C} \sum_{t=1}^{K}\sum_{s=t+1}^{K}\sqrt{2 - 2 {\mathbf{w}_{js}}^{\mathrm{T}} \mathbf{w}_{jt}}} {C K \left( K - 1 \right)}.
\end{equation}
This regularization encourages the proxies $\mathbf{w}_{ck}$ to get closer to the same class centers.
Finally, SoftTriple loss $L_{SoftTriple}$ is formulated according to \cite{SoftTriple}
\begin{equation}
	L_{SoftTriple} = \sum_{\mathbf{x} \in X} L_{Sim} \left( \mathbf{x} \right) + \tau L_{Regularizer}.
\end{equation}

SoftTriple loss may have the problems with learning process, such as ProxyNCA loss \cite{ProxyAnchor}.
Therefore, we investigate the gradient of SoftTriple loss, and we check the characteristics of SoftTriple loss.
The gradient of SoftTriple loss $L_{Sim}$ with respect to the similarity $\mathcal{S} \left( \mathbf{x}_{i}, c \right)$  is as follows
\begin{equation}
	\frac{\partial L_{Sim} \left( \mathbf{x}_{i} \right)}{\partial \mathcal{S} \left( \mathbf{x}_{i}, c \right)} =
	\begin{dcases}
		\frac{- \lambda}{\sum_{c^{\prime} \neq c_{i}} l_{R}^{+} \left( c_{i}, c^{\prime} \right) + 1}, & if \ c = c_{i},\\
		\frac{\lambda}{l_{R}^{+} \left( c_{i}, c \right) + \sum_{c^{\prime} \neq c_{i}} l_{R} \left( c, c^{\prime} \right)}, & \mathrm{otherwise},
	\end{dcases}
\end{equation}
where $l_{R}^{+} \left( c_{1}, c_{2} \right) = \exp \left( \lambda \left( S \left( \mathbf{x}_{i}, c_{1} \right) - S \left( \mathbf{x}_{i}, c_{2} \right) - \delta \right) \right)$ and $l_{R}^{-} \left( c_{1}, c_{2} \right) = \exp \left( \lambda \left( S \left( \mathbf{x}_{i}, c_{1} \right) -S \left( \mathbf{x}_{i}, c_{2} \right) \right) \right)$.
The gradient only considers the relations of similarities for class c and other classes.
Thus, this gradient cannot consider the magnitude of similarity $\mathcal{S} \left( \mathbf{x}_{i}, c \right)$ itself.
From these characteristics, the gradient could be small, even if the similarity itself is large enough for both positive and negative classes.

\subsection{Multi Proxies Anchor Loss}
We propose multi-proxies anchor (MPA) loss, using has multi-proxies and loss structures such as multi similarity (MS) loss \cite{MS}.
Thus, MPA loss is a proxy-based loss and could be considered as an extension of SoftTriple loss and ProxyAnchor loss \cite{SoftTriple,ProxyAnchor}.
Furthermore, multi-proxies are valid representations for real-world datasets, such as each class with multiple local centers
Additionally, the MS loss structure is also useful for problems with the gradient of softmax-based loss \cite{SoftTriple,ProxyAnchor}.

MPA loss computes the similarity between data and proxies using an inner product and the similarity between data and classes using softmax in the same way as \eqref{original_similarity}.
Note that, there are several ways to calculate similarity, such as the maximum and average similarity strategies.
MPA similarity loss is computed for all batch data and proxy center combinations.
MPA similarity loss $L_{MPASim}$ is defined as
\begin{align}
	\label{MPASim}
	L_{MPASim} &= \frac{1}{\left| C^{+} \right|} \sum_{c \in C^{+}} \log \left( 1 + \sum_{\mathbf{x} \in X_{c}^{+}} \exp \left( - \alpha \left( \mathcal{S} \left( \mathbf{x}, c \right) - \delta \right) \right) \right) \nonumber \\
	&+ \frac{1}{\left| C \right|} \sum_{c \in C} \log \left( 1 + \sum_{\mathbf{x} \in X_{c}^{-}} \exp \left( \alpha \left( \mathcal{S} \left( \mathbf{x}, c \right) + \delta \right) \right) \right),
\end{align}
where $C$ denotes the set of all classes, and $C^{+}$ denotes the set of positive classes in the batch data.
Also, $X_{c}^{+}$ denotes the positive data against class $c$; on the other hand, $X_{c}^{-}$ denotes the negative data against class $c$.
Note that, $\left| X_{c}^{+} \right| + \left| X_{c}^{-} \right|$ is equal to the batch size. 
This MPA similarity loss structure becomes the extension of MS loss and ProxyAnchor loss.
Finally, MPA loss combined with MPA similarity loss \eqref{MPASim} and proxies regularization \eqref{CenterReg}, and MPA loss is defined as
\begin{equation}
	\label{MPA}
	L_{MPA} = L_{MPASim} + \tau L_{Regularizer}.
\end{equation}

To compare the loss, we compare the gradient of MPA loss \eqref{MPA} and SoftTriple loss.
The gradient of MPA similarity loss \eqref{MPASim} with respect to the similarity $\mathcal{S} \left( \mathbf{x}, c \right)$ is derived according to
\begin{equation}
	\label{partial_MPA}
	\frac{\partial L_{MPASim}}{\partial \mathcal{S} \left( \mathbf{x}, c \right)} = 
	\begin{dcases}
		\frac{1}{\left| C^{+} \right|} \frac{- \alpha \ l_{MPA}^{+} \left( \mathbf{x}, c \right)}{1 + \sum_{\mathbf{x}^{\prime} \in X_{c}^{+}} l_{MPA}^{+} \left( \mathbf{x}^{\prime}, c \right)}, & \forall \mathbf{x} \in X_{c}^{+}, \\
		\frac{1}{\left| C \right|} \frac{\alpha \ l_{MPA}^{-} \left( \mathbf{x}, c \right)}{1 + \sum_{\mathbf{x}^{\prime} \in X_{c}^{-}} l_{MPA}^{-} \left( \mathbf{x}^{\prime}, c \right)}, & \forall \mathbf{x} \in X_{c}^{-}, 
	\end{dcases}
\end{equation}
where let $l_{MPA}^{+} \left( \mathbf{x}, c \right) = \exp \left( - \alpha \left( \mathcal{S} \left( \mathbf{x}, c \right) - \delta \right) \right)$ and $l_{MPA}^{-} \left( \mathbf{x}, c \right) = \exp \left( \alpha \left( \mathcal{S} \left( \mathbf{x}, c \right) + \delta \right) \right)$.
Additionally, this gradient~\eqref{partial_MPA} can be transformed as follows:
\begin{equation}
	\label{partial_MPA2}
	\frac{\partial L_{MPASim}}{\partial \mathcal{S} \left( \mathbf{x}, c \right)} = 
	\begin{dcases}
		\frac{1}{\left| C^{+} \right|} \frac{- \alpha}{l_{\mathcal{S}}^{+} + \sum_{\mathbf{x}^{\prime} \in X_{c}^{+}} \exp \left( -\alpha \mathcal{S}_{R} \right)}, & \forall \mathbf{x} \in X_{c}^{+}, \\
		\frac{1}{\left| C \right|} \frac{\alpha}{l_{\mathcal{S}}^{-} + \sum_{\mathbf{x}^{\prime} \in X_{c}^{-}} \exp \left( \alpha \mathcal{S}_{R} \right)}, & \forall \mathbf{x} \in X_{c}^{+}, \\
	\end{dcases}
\end{equation}
where let $l_{\mathcal{S}}^{+} = \exp \left( -\alpha \left( \delta - \mathcal{S} \left( \mathbf{x}, c\right) \right) \right)$, $l_{\mathcal{S}}^{-} = \exp \left( -\alpha \left( \mathcal{S} \left( \mathbf{x}, c\right) + \delta \right) \right)$, and $\mathcal{S}_{R} = \mathcal{S} \left( \mathbf{x}^{\prime}, c \right) - \mathcal{S} \left( \mathbf{x}, c \right)$.
The gradient of MPA loss occurs two factors each positive or negative class.
The first factor is the similarity loss between data $\mathbf{x}$ and class, such as $l_{\mathcal{S}}^{+}$ and $l_{\mathcal{S}}^{-}$.
The second factor is the loss of relative similarity between data $\mathbf{x}$ and the same class data, such as $\mathcal{S}_{R}$.
SoftTriple loss only considers the loss of relative similarities for anchor data $\mathbf{x}$.
Alternatively, MPA loss considers the loss of relative similarity for class and the similarity between data $\mathbf{x}$ and class.
Thus, compared to SoftTriple loss, MPA loss can consider the similarity scale itself, and MPA loss could lead to efficient learning and learning of complex manifold structures.

\subsection{Data-wise Multi-proxies Anchor Loss}
According to the gradient of MPA loss and ProxyAnchor loss~\eqref{partial_MPA2}, these losses lead to learning based on similarity itself and similarity relations in each class.
However, batch samples significantly affect this gradient because similarity relations consider a difference between $\mathbf{x}$ and $\mathbf{x}^{\prime}$ in the same positive or negative situation against the class.

Now, for MPA loss and ProxyAnchor loss structure~\eqref{MPASim}, this structure is similar to a soft-plus function.
Moreover, the soft-plus function with summation has an interesting characteristic.
For example, the function $f_{SoftPlus}$ of soft-plus with summation for instances $x_{i}$ is defined as
\begin{equation}
	\label{SoftPlus}
	f_{SoftPlus} = \log \left( 1 + \sum_{i}^{n} \exp x_{i} \right).
\end{equation}
The gradient of function $f_{SoftPlus}$ on the instance $x_{j}$ is derived as
\begin{equation}
	\frac{\partial f_{SoftPlus}}{\partial x_{j}} = \frac{1}{\exp \left( - x_{j} \right) + \sum_{i}^{n} \exp \left( x_{i} - x_{j} \right)}.
\end{equation}
Thus, the soft-plus function with summation has a gradient that considers an instance itself and relations between instances.

Similarly, MPA loss has different characteristics that changes the order of summation.
Therefore, we propose data-wise multi-proxies anchor (MPA-DW) loss for minimize the batch sample effect.
Changing the order of summation in \eqref{MPASim}, MPA-DW loss applies positive or negative similarity losses to the soft-plus function like \eqref{SoftPlus} each data.
Therefore, MPA-DW similarity loss $L_{MPADWSim}$ is defined as
\begin{align}
	\label{MPADWSim}
	L_{MPADWSim} &= \frac{1}{N} \sum_{i}^{N} \left( \log \left( 1 + \exp \left( - \alpha \left( \mathcal{S} \left( \mathbf{x}_{i}, c_{i} \right) - \delta \right) \right) \right) \right. \nonumber \\
	& \left. + \log \left( 1 + \sum_{c \neq c_{i}}\exp \left( \alpha \left( \mathcal{S} \left( \mathbf{x}_{i}, c \right) + \delta \right) \right) \right) \right).
\end{align}
Finally, MPA-DW loss combined with MPA-DW similarity loss \eqref{MPADWSim} and proxies regularization \eqref{CenterReg}, and MPA-DW loss $L_{MPADW}$ follows
\begin{equation}
	\label{MPADW}
	L_{MPADW} = L_{MPADWSim} + \tau L_{Regularizer}.
\end{equation}

The gradient of MPA-DW similarity loss \eqref{MPADWSim} to the similarity $\mathcal{S} \left( \mathbf{x}_{i}, c^{\prime} \right)$ is derived as
\begin{equation}
	\label{partial_MPADW}
	\frac{\partial L_{MPADWSim}}{\partial \mathcal{S} \left( \mathbf{x}_{i}, c^{\prime} \right)} = 
	\begin{dcases}
		\frac{1}{N} \frac{- \alpha}{l_{\mathcal{DS}}^{+} + 1}, & c^{\prime} = c_{i}, \\
		\frac{1}{N} \frac{\alpha}{l_{\mathcal{DS}}^{-} + \sum_{c \neq c_{i}} \exp \left( \alpha \mathcal{S}_{DR} \right)}, & c^{\prime} \neq c_{i}, \\
	\end{dcases}
\end{equation}
where let $l_{\mathcal{DS}}^{+} = \exp \left( - \alpha \left( \delta - \mathcal{S} \left( \mathbf{x}_{i}, c_{i} \right) \right) \right)$, $l_{\mathcal{DS}}^{-} = \exp \left( - \alpha \left( \mathcal{S} \left( \mathbf{x}_{i}, c^{\prime} \right) + \delta \right) \right)$, and $\mathcal{S}_{DR} = \mathcal{S} \left( \mathbf{x}_{i}, c^{\prime} \right) - \mathbf{S} \left( \mathbf{x}_{i} c^{\prime} \right)$.
This gradient also considers the similarity itself and similarity relations. 
Additionally, this gradient removes batch sample effects since similarity relations are the difference between similarities from the same data in each class.
However, a positive gradient cannot consider similarity relations because datasets are generally one class for one data.

\begin{figure*}[tb]
	\begin{center}
		\includegraphics[width=\hsize]{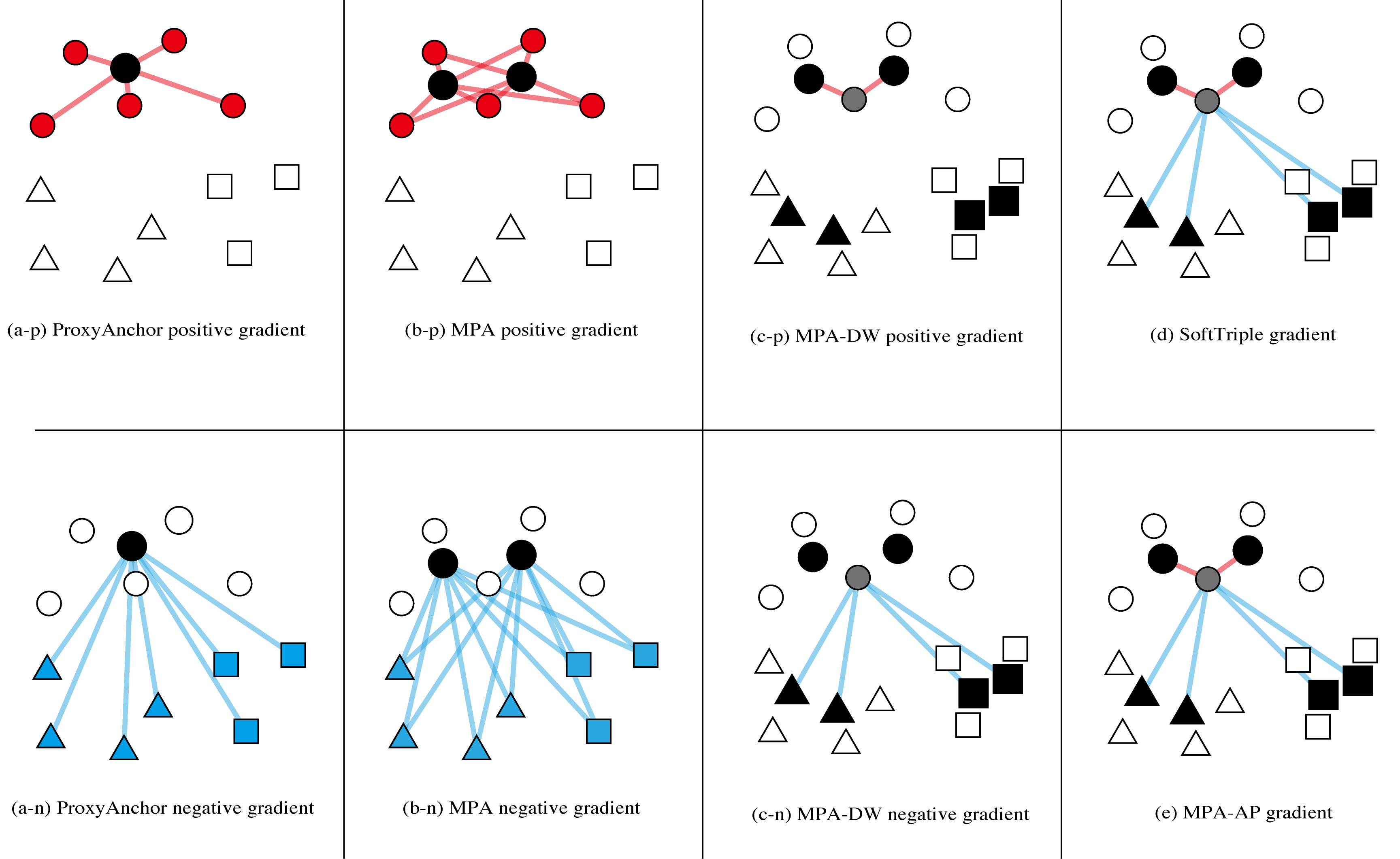}
	\end{center}
	\caption{The comparison of the gradients of ProxyAnchor loss, SoftTriple loss, MPA loss, MPA-DW loss, and MPA-AP loss.
	Each figure illustrates the similarities affecting the respective gradients.
	Symbol shape denotes classes.
	Red symbols represent the positive embeddings, blue symbols represent the negative embeddings, gray symbols represent the anchor embeddings, and black symbols represent the proxy centers.
	These losses cause gradients so that similar data are near (red lines are short) and dissimilar data are far (blue lines are long).
	ProxyAnchor loss, MPA loss, and MPA-DW loss have two types of gradients for positive and negative, and SoftTriple loss and MPA-AP loss have only one type.
	In addition, ProxyAnchor loss and MPA loss focus on the black circle symbol (proxies), and SoftTriple loss, MPA-DW loss, and MPA-AP loss focus on the gray circle symbol (anchor embedding).
	}
	\label{DifferenceLosses}
\end{figure*}

This problem arises because both MPA loss and MPA-DW loss separate the positive and negative similarity losses.
Therefore, we also propose the data-wise and all-paired multi-proxies anchor (MPA-AP) loss.
In the first step, we redefine the similarity notation.
We introduce the similarity with margin as follows
\begin{equation}
	\mathcal{S}^{\prime} \left( \mathbf{x}, c \right) = 
	\begin{dcases}
		\delta - \mathcal{S} \left( \mathbf{x}, c \right), & c = c_{i}, \\
		\mathcal{S} \left( \mathbf{x}, c \right) + \delta, & c \neq c_{i}. \\
	\end{dcases}
\end{equation}
Then, MPA-AP's objective of loss functions is to minimize this similarities $\mathcal{S}^{\prime} \left( \mathbf{x}, c \right)$.
MPA-AP loss considers all pair similarity relations regardless of positive and negative similarities.
MPA-AP loss computes the loss by changing the order of summation in the same way as MPA-DW loss.
First, MPA-AP similarity loss $L_{MPAAPSim}$ is defined as
\begin{equation}
	L_{MPAAPSim} = \frac{1}{N} \sum_{i}^{N} \log \left( 1 + \sum_{c} \exp \left( \alpha \mathcal{S}^{\prime} \left( \mathbf{x}_{i}, c \right) \right) \right).
\end{equation}
Then, MPA-AP loss $L_{MPAAP}$ is formulated following
\begin{equation}
	\label{MPAAP}
	L_{MPAAP} = L_{MPAAPSim} + \tau \L_{Regularizer}.
\end{equation}
Finally, the gradient of MPA-AP similarity loss is derived as
\begin{equation}
	\label{MPAAP_gradient}
	\frac{\partial L_{MPAAPSim}}{\partial \mathcal{S}^{\prime} \left( \mathbf{x}_{i}, c^{\prime} \right)} = \frac{1}{N} \frac{\alpha}{\exp \left( - \alpha \mathcal{S}^{\prime} \left( \mathbf{x}_{i}, c^{\prime} \right) \right) + \sum_{c} \left( \exp \left( \alpha \mathcal{S}^{\prime}_{R} \right) \right)},
\end{equation}
where let $\mathcal{S}^{\prime}_{R} = \mathcal{S}^{\prime} \left( \mathbf{x}_{i}, c \right) - \mathcal{S}^{\prime} \left( \mathbf{x}_{i}, c^{\prime} \right)$.
This gradient \eqref{MPAAP_gradient} can consider the similarity itself and all relations of similarities between classes for each data.
Compared to MPA-DW loss, MPA-AP loss can consider similarity relations between positive and negative classes.
Additionally, MPA-AP loss mitigates the batch sample effect as well as MPA-DW loss compared to MPA loss.
Therefore, MPA-AP loss has advantages compared with MPA loss, MPA-DW loss, and SoftTriple loss.
Figure \ref{DifferenceLosses} shows the differences in proxy-based loss gradients.
ProxyAnchor loss and MPA loss have occurred in gradients centered on proxies.
Alternatively, SoftTriple loss, MPA-DW loss, and MPA-AP loss have occurred in data-centered gradients.
In summary, the characteristics of proxy-based losses are shown in table \ref{GradientComp}.
Table \ref{GradientComp} shows that SoftTriple loss, MPA-DW loss, and MPA-AP loss have high stability.
Additionally, ProxyNCA loss and MPA loss, can consider relations between positive similarities instead of stability.
\begin{table*}[tb]
	\caption{The comparison of the proxy-based loss characteristics. This table shows the evaluation from six perspectives. Hereinafter, positive similarity means the similarity between data and positive class, and negative similarity means the similarity between data and negative class. The first perspective, stability, means the enormity of batch samples on the gradients. Then, manifold means losses can consider various class distributions or not. The other views show factors of gradients on DML's losses. Similarity means the size of positive or negative similarity. P-P relations show differences of both positive similarities. Then, N-N relations also show differences of both negative similarities. Finally, P-N relations show differences between positive similarity and negative similarity. (\checkmark) means that it is not necessarily taken into account.}
	\centering{
		\begin{tabular}{c|c|c|c|c|c|c}
			\hline
			Loss & Stablity & Manifold & Similarity & P-P relations & N-N relations & P-N relations \\
			\hline
			ProxyNCA \cite{ProxyNCA} & \checkmark & \ & \ & \ & \checkmark & \  \\
			ProxyAnchor \cite{ProxyAnchor} & \ & \ & \checkmark & (\checkmark) & \checkmark & \ \\
			SoftTriple \cite{SoftTriple} &\checkmark & \checkmark & \  & \  & \checkmark & \checkmark \\
			\hline
			MPA & \  & \checkmark & \checkmark & (\checkmark) & \checkmark & \  \\
			MPA-DW & \checkmark & \checkmark & \checkmark & \  & \checkmark & \  \\
			MPA-AP & \checkmark & \checkmark & \checkmark & \  & \checkmark & \checkmark \\
			\hline
		\end{tabular}
		\label{GradientComp}
	}
\end{table*}

\subsection{Training Complexity}
The training complexity depends on the loss types because loss types change the number of combinations.
M, C, and P describe the number of training data, classes, and proxies, respectively.
These numbers are generally $M > C > P$.
The comparison of training complexity is shown in table \ref{TrainingComplexity}.
Pair-based losses typically have higher training complexity than proxy-based losses because they consider combinations between training samples.
The training complexity of the contrastive loss is $O ( M^{2} )$ because it computes a pair of samples
Additionally, triplet loss and MultiSimilarity loss are $O ( M^{3} )$ because these losses sample triplets of data and compare positive pair (same class data pair) to multi-negative pair (different class data pair), respectively.

Because proxy-based losses compute combinations between training data and proxies, these losses have lower training complexity.
Additionally, the complexity of proxy-based losses complexity greatly depends on the number of proxies.
The complexity of ProxyNCA loss and ProxyAnchor loss, which have a single-proxy for each class, are both $O ( M C )$.
The complexity of SoftTriple loss is $O ( M C P^{2} )$ since this loss compares combinations for each data with $P$ positive proxies and $C (P-1)$ negative proxies.

MPA loss, MPA-DW loss, and MPA-AP loss have the same complexity as proxy-based losses.
First, the complexity of MPA loss is $O ( M C P^{2} )$ because it compares each proxy to all positive or negative samples in a batch.
Second, MPA-DW loss associates each data with positive or negative proxies.
Therefore, its complexity is $O ( M C P^{2} )$.
Finally, the complexity of MPA-AP loss is also $O ( M C P^{2} )$ because it associates each data with all proxies.
Thus, MPA-family losses increase complexity more than single-proxy losses. 
\begin{table*}[tb]
	\caption{The comparison of training complexity.}
	\centering{
		\begin{tabular}{c|c|c}
			\hline
			Type & Loss & Training Complexity \\
			\hline
			\ & Contrastive \cite{Contrastive} & $O ( M^{2} )$ \\
			Pair-based & Triplet \cite{Triplet} & $O ( M^{3} )$ \\
			\ & MultiSimilarity \cite{MS} & $O ( M^{3} )$ \\
			\hline
			\ & ProxyNCA \cite{ProxyNCA} & $O ( MC )$  \\
			Proxy-based & ProxyAnchor \cite{ProxyAnchor} & $O ( MC )$ \\
			\ & SoftTriple \cite{SoftTriple} & $O ( MCP^{2} )$ \\
			\hline
			\ & MPA & $O ( MCP^{2} )$ \\
			Proxy-based & MPA-DW & $O ( MCP^{2} )$ \\
			\ & MPA-AP & $O ( MCP^{2} )$ \\
			\hline
		\end{tabular}
		\label{TrainingComplexity}
	}
\end{table*}

\subsection{Effective Deep Metric Learning Performance Metric}
Traditionally, Recall@k and Normalized Mutual Information (NMI) metrics are used as the DML performance metrics \cite{LiftedStructure, NMI, MS, SoftTriple,ProxyAnchor}.
However, these metrics cannot fully evaluate the DML performance, which is image retrieval performance \cite{RealityCheck}.
Then, a new performance metric, MAP@R, which combines mean average precision and R-precision, is proposed.
MAP@R is more stable than recall@k and NMI as DML performance metrics \cite{RealityCheck}.
When evaluating the DML performance, the search length R of MAP@R is the number of positive data in the datasets.
Note that in MAP@R, the search length is different for each class.
In the DML datasets, the size of the positive data for each class is much smaller or larger. 
When the positive data size is small, MAP@R cannot evaluate the DML performance sufficiently due to the short search length.  
Additionally, assuming that each class has some local centers, MAP@R cannot evaluate the DML performance sufficiently when the size of positive data is large.
Therefore, the DML performance evaluation must  validate some search length.
Note that MAP@k, where k is any search length, may fail to evaluate the DML if the search length is longer than the number of positive class data.

We propose the nDCG@k metric as a DML performance metric, where nDCG is conventionally used as an evaluation metric for the rank function in search engines \cite{NDCG1, NDCG2}.
nDCG@k has better stability than the Recall@k metric and more flexibility than the MAP@R metric.
nDCG@k is computed using the discounted cumulative gain (DCG) of search results and the DCG of the best search results.
In this paper, the DCG is defined as
\begin{equation}
	DCG \left( k \right) = \sum_{i=1}^{k} \frac{2^{r_{i}} - 1}{\log_{2} \left( i + 1 \right)},
\end{equation}
where $r_{i}$ denotes the $i$th search result score and $k$ denotes the search length.
$r_{i}$ is a binary value when evaluating the DML performance.
$r_{i}=1$ indicates that the $i$th search result is positive class data for the query data, while $r_{i}=0$ indicates the opposite result.
Finally, the nDCG@k metric is defined as
\begin{equation}
	nDCG \left( k \right) = \frac{DCG \left( k \right)}{DCG_{best} \left( k \right)},
\end{equation}
where $DCG_{best} \left( k \right)$ denotes the DCG when the best search results.
In this paper, we deal with 100 times nDCG@k to align with conventional performance metrics.

To compare the DML performance metrics, table \ref{MetricsExamples} shows examples when the number of positive data against the query is four.
According to table \ref{MetricsExamples}, Recall@10, Precision@10, MAP@R, and MAP@10 cannot evaluate the search results sufficiently, while nDCG@10 can evaluate the performance sufficiently in all results.
\begin{table*}[tbp]
	\caption{The comparison of the DML performance metrics. The DML performance metrics show Recall@10, Precision@10, MAP@R where R is four, and nDCG@10. The search results are in the order from left to right. Last column result shows the best search result.}
	\centering{
		\begin{tabular}{c|c|c|c|c|c}
			\hline
			Search results & Recall@10 & Precision@10 & MAP@R & MAP@10 & nDCG@10 \\
			\hline
			[1, 0, 0, 0, 0, 0, 0, 0, 0, 0] & 100 & 10 & 25.0 & 10.0 & 39.0 \\
			\hline
			[1, 0, 0, 0, 0, 0, 0, 0, 0, 1] & 100 & 20 & 25.0 & 12.0 & 50.3 \\
			\hline
			[1, 0, 1, 0, 0, 0, 0, 0, 0, 0] & 100 & 20 & 41.7 & 16.7 & 58.6 \\
			\hline
			[1, 0, 1, 0, 0, 0, 1, 0, 0, 1] & 100 & 40 & 41.7 & 25.0 & 82.9 \\
			\hline
			[1, 1, 1, 1, 0, 0, 0, 0, 0, 0] & 100 & 40 & 100.0 & 40.0 & 100.0 \\
			\hline
		\end{tabular}
		\label{MetricsExamples}
	}
\end{table*}

\section{Evaluation}
In this section, we compare MPA-family losses with state-of-the-art losses for the effectiveness of loss.
First, we evaluate the loss performance on three benchmark datasets for image retrieval and fine-grained tasks.
Then, we also validate the impact of the number of proxies.

\subsection{Datasets}
We evaluate the proposed losses on three datasets. 
The datasets are widely-used benchmark datasets, fine-grained datasets, and large-scale few-shot image datasets.
These datasets are CUB-200-2011 \cite{CUB}, Cars196 \cite{Cars}, and Stanford Online Products \cite{LiftedStructure} dataset.

CUB-200-2011 \cite{CUB} contains 11,788 bird images in 200 classes.
We split the dataset into two so that the number of classes is even.
Following, we use 5,924 images with 100 classes for the training and 5,864 images with 100 classes for the test.
Cars196 \cite{Cars} contains 16,185 car images with 196 classes.
Similar to CUB-200-2011, we split the dataset into two so that the number of classes is even.
The training and test data are 8,054 images in 98 classes and 8,131 images, respectively.
Stanford Online Products (SOP) contains 120,053 images in 22,634 categories.
This dataset uses 59,551 images in 11,318 categories for training and 60,502 images in 11,316 categories for the test.

\subsection{Implementation Details}
We use the Inception \cite{Inception} with batch normalization \cite{BN} (InceptionBN), as in the previous studies \cite{SoftTriple, MS, ProxyAnchor}.
The parameters are trained on the ImageNet ILSVRC 2012 dataset \cite{ImageNet}.
Then, these parameters are fine-tuned on the target dataset \cite{finetune}.
The output vectors have 512 dimensions in our experiments.
We apply the cubic GeM \cite{GeM} as a global descriptor to the outputs of the backbone network.
In the data preprocessing setting, both training and test images are resized to any size and cropped to $224 \times 224$.
Training images are randomly resized and cropped to $224 \times 224$, then randomly flipped horizontally for data augmentation.
Alternatively, test images are resized to $256 \times 256$ and cropped to $224 \times 224$ at the image center.
The optimization method uses the AdamW optimizer \cite{AdamW} for all experiments.
The initial learning rates for the backbone and proxies are $1.0 \times 10^{-4}$ and $1.0 \times 10^{-2}$, respectively.
The training batch sizes for CUB-200-2011, Cars196, and SOP are 180, 64, and 32, respectively.
The number of epochs is 60 for CUB-200-2011 and Cars196, and the number of epochs is 100 for SOP.
Batch sampling uses random sampling, the same as SoftTriple loss \cite{SoftTriple}.
We decay these learning rates by $0.5$ every 20 epochs in all datasets.
We set $\tau = 0.2$, $\gamma=0.1$, and $\delta = 0.1$ in \eqref{MPA}, \eqref{MPADW}, and \eqref{MPAAP}.
The number of centers is set to $K=2$ for SOP and $K=10$ for CUB-200-2011 and Cars196.
For fairness, we experiment with the SoftTriple loss and ProxyAnchor loss in the same architecture and settings and compare them with MPA loss.
Additionally, MPA-family loss, ProxyAnchor loss, and SoftTriple loss are learned three times with the same settings.

\subsection{Comparison of state-of-the-art Losses}
In this section, we compared MPA loss with state-of-the-art losses.
We show the comparison of recall@k and nDCG@k on CUB-200-2011, Cars196, and SOP in table \ref{CUB_Result}, table \ref{Cars_Result}, and table \ref{SOP_Result}, respectively.
In CUB-200-2011 and Cars196, MPA loss and MPA-AP loss had the best performance compared to the other DML losses.
Furthermore, MPA loss and MPA-AP loss were mostly the same accuracies in the two datasets.
Alternatively, SoftTriple loss was the best performance compared to MPA-family losses and ProxyAnchor loss, and MPA loss and ProxyAnchor loss accuracies were not much different in the SOP dataset.

In terms of MAP@R, unlike the other metrics, ProxyAnchor loss was the highest MAP@R.
These differences would be based on the search length.
Search length R depends on the number of positive data each class.
The mean of R is $58.64$, $84.70$, and $5.35$ in CUB-200-2011, Cars196, and SOP, respectively.
The search length R is too long for CUB-200-2011 and Cars196 and too short for SOP.
Generally, as the amount of data per class increases, each class distribution forms a manifold, making an evaluation with long search lengths inappropriate.
Alternatively, MAP@R lacks flexibility and comparability for short search lengths, as shown in table \ref{MetricsExamples}.
Therefore, our experiment mainly uses Recall@1 and nDCG@k.

In CUB 200-2011 and Cars196, MPA-family losses improved accuracy.
Alternatively, in SOP, MPA-family losses did not improve the accuracy.
These results may be related to the number of classes and the mean of positive data for each class in the training datasets.
The mean of positive data for each class is $59.2$, $82.2$, and $5.3$ in CUB-200-2011, Cars196, and SOP, respectively.
This difference is expected to affect the result for MPA-family losses and ProxyAnchor loss.
MPA-family losses are superior to ProxyAnchor loss when the mean of positive data is large, such as CUB-200-2011 and Cars196, while MPA loss is mostly unchanged from ProxyAnchor loss when the mean of positive data is small, such as SOP.
If the mean is large, each class could have several local centers, as assumed in  MPA-family losses.
Conversely, if the mean is small, each class might have only a center.
Therefore, MPA-family losses are more accurate in CUB-200-2011 and Cars196, and MPA loss and ProxyAnchor loss are almost the same accuracies in SOP.

Furthermore, all results will be related to the characteristics presented in table~\ref{GradientComp}.
Particularly, the results at SOP reflect the characteristics:
the difference in accuracy between MPA-AP loss and MPA loss may be due to the mitigation of sampling effects in an increasing number of combinations.
On the other hand, the difference in accuracy between MPA-family loss and SoftTriple loss may be due to the influence of the similarity itself.

Among MPA-family losses, MPA-AP loss was as accurate or more accurate than the other losses in all datasets.
The most significant difference between MPA-AP loss and MPA loss is whether it is a data-wise loss or class-wise loss.
Data-wise loss is less sensitive to the sampling effect than class-wise loss.
Hence, MPA-AP loss was more accurate for SOP, which has more sampling combinations.
Alternatively, MPA-AP loss can learn to consider most of the factors shown in table~\ref{GradientComp}.
Therefore, MPA-AP loss is superior to the other MPA-family losses.

\begin{table*}[tbp]
	\caption{This table shows the comparison of recall@k, nDCG@k, and MAP@R on CUB-200-2011. Bold style values represent the highest accuracy, and the value in parentheses represents the standard deviation of accuracy. (our) means the result of a rerun of the same experimental setup as this experiment.}
	\centering{
		\begin{tabular}{c|cccc|ccc|c}
			\hline
			Methods & R@1 & R@2 & R@4 & R@8 & nDCG@2 & nDCG@4 & nDCG@8 & MAP@R \\
			\hline
			Clusteing \cite{Clustering} & $48.2$ & $61.4$ & $71.8$ & $81.9$ & - & - & - & - \\
			ProxyNCA \cite{ProxyNCA} & $49.2$ & $61.9$ & $67.9$ & $72.4$ & - & - & - & - \\
			HDC \cite{HDC} & $53.6$ & $65.7$ & $77.0$ & $85.6$ & - & - & - & - \\
			Margin \cite{Margin} & $63.6$ & $74.4$ & $83.1$ & $90.0$ & - & - & - & - \\
			HTL \cite{HTL} & $57.1$ & $68.8$ & $78.7$ & $86.5$ & - & - & - & - \\
			MS \cite{MS} & $65.7$ & $77.0$ & $86.3$ & $91.2$ & - & - & - & - \\
			SoftTriple \cite{SoftTriple} & $65.4$ & $76.4$ & $84.5$ & $90.4$ & - & - & - & - \\
			ProxyAnchor \cite{ProxyAnchor} & $68.4$ & $79.2$ & $86.8$ & $91.6$ & - & - & - & - \\
			\hline
			SoftTriple (our) & $67.6$ & $78.0$ & $85.8$ & $91.1$ & $65.4$ & $62.9$ & $59.4$ & $25.5$ \\
			ProxyAnchor (our) & $68.6$ & $78.7$ & $86.1$ & $91.7$ & $66.7$ & $64.1$ & $60.9$ & $\mathbf{26.7}$ \\
			MPA & $\mathbf{69.2}$ & $\mathbf{79.4}$ & $\mathbf{86.6}$ & $91.6$ & $\mathbf{67.3}$ & $\mathbf{64.7}$ & $\mathbf{61.3}$ & $26.6$ \\
			MPA-DW & $68.9$ & $79.0$ & $86.5$ & $\mathbf{91.9}$ & $66.9$ & $64.3$ & $61.1$ & $26.3$ \\
			MPA-AP & $69.1$ & $79.1$ & $86.3$ & $91.7$ & $67.1$ & $64.6$ & $\mathbf{61.3}$ & $26.6$ \\
			\hline
		\end{tabular}
		\label{CUB_Result}
	}
\end{table*}

\begin{table*}[tbp]
	\caption{This table shows the comparison of recall@k and nDCG@k on Cars196. Bold style values represent the highest accuracy, and the value in parentheses represents the standard deviation of accuracy. (our) means the result of a rerun of the same experimental setup as this experiment.}
	\centering{
		\begin{tabular}{c|cccc|ccc|c}
			\hline
			Methods & R@1 & R@2 & R@4 & R@8 & nDCG@2 & nDCG@4 & nDCG@8 & MAP@R \\
			\hline
			Clusteing \cite{Clustering} & $58.1$ & $70.6$ & $80.3$ & $87.8$ & - & - & - & - \\
			ProxyNCA \cite{ProxyNCA} & $73.2$ & $82.4$ & $86.4$ & $88.7$ & - & - & - & - \\
			HDC \cite{HDC} & $73.7$ & $83.2$ & $89.5$ & $93.8$ & - & - & - & - \\
			Margin \cite{Margin} & $79.6$ & $86.5$ & $91.9$ & $95.1$ & - & - & - & - \\
			HTL \cite{HTL} & $81.4$ & $88.0$ & $92.7$ & $95.7$ & - & - & - & - \\
			MS \cite{MS} & $84.1$ & $90.4$ & $94.0$ & $96.5$ & - & - & - & - \\
			SoftTriple \cite{SoftTriple} & $84.5$ & $90.7$ & $94.5$ & $96.9$ & - & - & - & - \\
			ProxyAnchor \cite{ProxyAnchor} & $86.1$ & $91.7$ & $95.0$ & $97.3$ & - & - & - & - \\
			\hline
			SoftTriple (our) & $86.0$ & $91.8$ & $95.2$ & $97.2$ & $84.1$ & $81.3$ & $77.1$ & $27.5$ \\
			ProxyAnchor (our) & $86.1$ & $91.6$ & $95.0$ & $97.2$ & $84.3$ & $81.6$ & $77.6$ & $\mathbf{28.8}$ \\
			MPA & $86.8$ & $92.0$ & $95.1$ & $97.2$ & $84.9$ & $82.1$ & $78.0$ & $28.4$ \\
			MPA-DW & $86.6$ & $91.8$ & $95.2$ & $97.2$ & $84.8$ & $82.1$ & $78.0$ & $28.1$ \\
			MPA-AP & $\mathbf{87.1}$ & $\mathbf{92.4}$ & $\mathbf{95.5}$ & $\mathbf{97.5}$ & $\mathbf{85.3}$ & $\mathbf{82.7}$ & $\mathbf{78.8}$ & $28.6$ \\
			\hline
		\end{tabular}
		\label{Cars_Result}
	}
\end{table*}

\begin{table*}[tbp]
	\caption{This table shows the comparison of recall@k and nDCG@k on SOP. Bold style values represent the highest accuracy, and the value in parentheses represents the standard deviation of accuracy. (our) means the result of a rerun of the same experimental setup as this experiment.}
	\centering{
		\begin{tabular}{c|ccc|cc|c}
			\hline
			Methods & R@1 & R@10 & R@100 & nDCG@10 & nDCG@100 & MAP@R \\
			\hline
			HDC \cite{HDC} & $69.5$ & $84.4$ & $92.8$ & - & - & - \\
			HTL \cite{HTL} & $74.8$ & $88.3$ & $94.8$ & - & - & - \\
			MS \cite{MS} & $78.2$ & $90.5$ & $96.0$ & - & - & - \\
			SoftTriple \cite{SoftTriple} & $78.3$ & $90.3$ & $95.9$  & - & - & - \\
			ProxyAnchor \cite{ProxyAnchor} & $79.1$ & $90.8$ & $96.2$ & - & - & - \\
			\hline
			SoftTriple (our) & $\mathbf{79.0}$ & $\mathbf{90.7}$ & $\mathbf{96.0}$ & $\mathbf{64.4}$ & $\mathbf{70.7}$ & $\mathbf{51.7}$ \\
			ProxyAnchor (our) & $77.6$ & $90.2$ & $95.9$ & $62.7$ & $69.2$ & $49.7$ \\
			MPA & $77.5$ & $90.0$ & $95.8$ & $62.5$ & $69.0$ & $49.6$ \\
			MPA-DW & $78.4$ & $90.1$ & $95.5$ & $63.2$ & $69.2$ & $50.5$ \\
			MPA-AP & $78.1$ & $90.1$ & $95.6$ & $62.9$ & $69.1$ & $50.2$ \\
			\hline
		\end{tabular}
		\label{SOP_Result}
	}
\end{table*}

\subsection{The Impacts of Proxies}
In this section, we also experiment with the effect of the number of proxies on MPA loss.
We evaluate the accuracy using the R@1 and nDCG@k metrics while varying the number of proxies $K$ on CUB-200-2011 and Cars196.
Note that this evaluation is the same implementation as the state-of-the-art loss comparison, except for the number of proxies.
This evaluation uses $K = \{1, 4, 8, 12, 16\}$, and MPA loss learns three times for each condition.

\begin{figure}[tbp]
	\begin{center}
		\includegraphics[width=0.7\hsize]{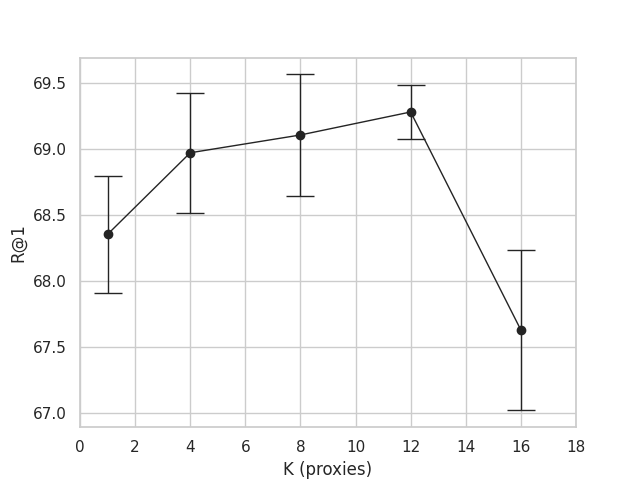}
	\end{center}
	\caption{R@1 metric results on CUB-200-2011 dataset while changing the number of proxies $K$.}
	\label{CUBR1Graph}
\end{figure}

\begin{figure}[tbp]
	\begin{center}
		\includegraphics[width=0.7\hsize]{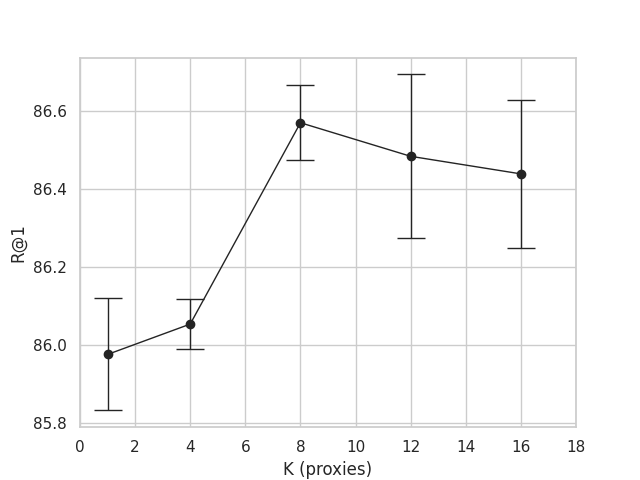}
	\end{center}
	\caption{R@1 metric results on Cars196 dataset while changing the number of proxies $K$.}
	\label{CarsR1Graph}
\end{figure}

\begin{table*}[tbp]
	\caption{This table shows the comparison of recall@1 and nDCG@k on CUB-200-2011. Bold style values represent the highest accuracy, and the value in parentheses represents the standard deviation of accuracy.}
	\centering{
		\begin{tabular}{c|cccc}
			\hline
			Proxies & R@1 & nDCG@2 & nDCG@4 & nDCG@8 \\
			\hline
			$K=1$ & $68.4$ & $66.7$ & $64.0$ & $60.9$ \\
			$K=4$ & $69.0$ & $67.2$ & $64.5$ & $61.2$ \\
			$K=8$ & $69.1$ & $\mathbf{67.4}$ & $64.7$ & $61.4$ \\
			$K=12$ & $\mathbf{69.3}$ & $67.3$ & $\mathbf{64.8}$ & $\mathbf{61.5}$ \\
			$K=16$ & $67.6$ & $66.1$ & $63.5$ & $60.2$ \\
			\hline
		\end{tabular}
		\label{CUB_ProxiesImpacts}
	}
\end{table*}
\begin{table*}[tbp]
	\caption{This table shows the comparison of recall@1 and nDCG@k on Cars196. Bold style values represent the highest accuracy, and the value in parentheses represents the standard deviation of accuracy.}
	\centering{
		\begin{tabular}{c|cccc}
			\hline
			Proxies & R@1 & nDCG@2 & nDCG@4 & nDCG@8 \\
			\hline
			$K=1$ & $86.0$ & $84.3$ & $81.4$ & $77.3$ \\
			$K=4$ & $86.1$ & $84.5$ & $81.7$ & $77.6$ \\
			$K=8$ & $\mathbf{86.6}$ & $\mathbf{84.9}$ & $\mathbf{82.1}$ & $\mathbf{78.1}$ \\
			$K=12$ & $86.5$ & $84.8$ & $\mathbf{82.1}$ & $78.0$ \\
			$K=16$ & $86.4$ & $84.7$ & $\mathbf{82.1}$ & $78.0$ \\
			\hline
		\end{tabular}
		\label{Cars_ProxiesImpacts}
	}
\end{table*}

Figure~\ref{CUBR1Graph} and \ref{CarsR1Graph} show the R@1 results as $K$ is varied on CUB-200-2011 and Cars196, respectively.
Additionally, table~\ref{CUB_ProxiesImpacts} and table~\ref{Cars_ProxiesImpacts} show the R@1 and nDCG@k metrics on CUB-200-2011 and Cars196, respectively.
According to these results, MPA loss had the highest accuracy when the number of proxies $K$ was equal to 8 or 12.
Alternatively, fewer proxies $K=1$ and $K=4$ or many proxies $K=16$ conditions were less accurate than $K=8$ or $K=12$.

From these results, MPA loss could perform better when set to the appropriate number of proxies on target datasets.
Thus, MPA loss with appropriate proxies could promote more flexible learning for practical datasets than single-proxy loss.
Furthermore, this truly appropriate number of proxies may correspond to the number of local centers in each class on the target dataset.
Note that these local centers arise from the intra-class variance, and these local centers for each class are difficult to know in advance.
Therefore, MPA loss must decide the number of proxies that is sufficient to properly represent local centers and proxies regularization, such as \eqref{CenterReg}.

\section{Conclusion}
We have proposed the multi-proxies anchor (MPA) family losses, which are extended SoftTriple loss and ProxyAnchor loss.
MPA loss has solved two problems of SoftTriple loss and ProxyAnchor loss.
(1) MPA loss is flexible, fitting real-world datasets with multiple local centers.
(2) MPA loss solves the gradient problems for backpropagation compared with SoftTriple loss, leading to higher accuracy.
Additionally, we also proposed data-wise multi-proxies anchor (MPA-DW) loss and all-paired multi-proxies anchor (MPA-AP) loss.
These losses can generate more stable gradients than MPA loss.
Particularly, MPA-AP was the most accurate of the MPA-family losses.
Therefore, data-wise losses are considered useful.
We have also proposed the normalized discounted cumulative gain (nDCG@k) metric as the effective DML performance metric.
The nDCG@k metric showed more flexibility and effectiveness while maintaining good stability than traditional DML performance metrics such as recall@k and MAP@R metrics.
Several factors influence the similarity, including object type, situation, and background.
Conventional DML approaches focus on the kind of objecttype, but our future work studies a new DML approach that considers multiple factors.
\section*{Acknowledgments}
This work was supported by JSPS KAKENHI Grant Number JP18K11528.

\bibliographystyle{elsarticle-num}
\bibliography{references}
\end{document}